\documentclass[a4paper,conference]{IEEEtran}
\usepackage[T1]{fontenc}
\usepackage[utf8]{inputenc}
\usepackage{amsmath,amssymb,amsfonts}
\usepackage{graphicx}
\usepackage{booktabs}
\usepackage{url}
\usepackage{hyperref}
\usepackage{cite}
\usepackage{float}
\usepackage{caption}
\usepackage{subcaption}
\usepackage{microtype}
\usepackage{siunitx}
\usepackage{courier}
\usepackage{tikz}

\newcommand{\calN}{\mathcal{N}}
\newcommand{\calC}{\mathcal{C}}
\newcommand{\calH}{\mathcal{H}}
\newcommand{\RR}{\mathbb{R}}
\newcommand{\eps}{\varepsilon}

\title{NeuFACO: Neural Focused Ant Colony Optimization for Traveling Salesman Problem}
\IEEEoverridecommandlockouts
\author{%
  \IEEEauthorblockN{ 
    Dat Thanh Tran\IEEEauthorrefmark{1}\IEEEauthorrefmark{3},
    Khai Quang Tran\IEEEauthorrefmark{1}\textsuperscript{1}, 
    Khoi Anh Pham\IEEEauthorrefmark{2}\textsuperscript{1},
    Van Khu Vu\IEEEauthorrefmark{1},
    Dong Duc Do\IEEEauthorrefmark{3}}
  \IEEEauthorblockA{\IEEEauthorrefmark{1}%
                    College of Engineering and Computer Science, VinUniversity, Hanoi, Vietnam\\
                    \IEEEauthorrefmark{2}%
                    School of Information and Communications Technology, Hanoi University of Science and Technology, Hanoi, Vietnam\\
                    \IEEEauthorrefmark{3}%
                    University of Engineering and Technology, Vietnam National University, Hanoi, Vietnam\\
                    }                
  \IEEEauthorblockA{\IEEEauthorrefmark{1}\{dat.tt3,23khai.tq,khu.vv\}@vinuni.edu.vn \quad
                     \IEEEauthorrefmark{2}khoi.pa230043@sis.hust.edu.vn \quad
                     \IEEEauthorrefmark{3}dongdoduc@vnu.edu.vn
                     }
                     \thanks{\textsuperscript{1}Equal contribution.}
}

\begin{document}

\maketitle

\begin{abstract}This study presents Neural Focused Ant Colony Optimization (NeuFACO), a non-autoregressive framework for the Traveling Salesman Problem (TSP) that combines advanced reinforcement learning with enhanced Ant Colony Optimization (ACO). NeuFACO employs Proximal Policy Optimization (PPO) with entropy regularization to train a graph neural network for instance-specific heuristic guidance, which is integrated into an optimized ACO framework featuring candidate lists, restricted tour refinement, and scalable local search. By leveraging amortized inference alongside ACO’s stochastic exploration, NeuFACO efficiently produces high-quality solutions across diverse TSP instances. 
\end{abstract}

\begin{IEEEkeywords}
Neural Combinatorial Optimization, Traveling Salesman Problem, Ant Colony Optimization, Proximal Policy Optimization, Graph Neural Networks
\end{IEEEkeywords}

\section{Introduction}
The Traveling Salesman Problem (TSP) is a classical combinatorial optimization problem (COP) that seeks the shortest Hamiltonian tour visiting each city once and returning to the start. Due to its relevance in logistics, production planning, and circuit design, it serves as a benchmark for evaluating optimization algorithms \cite{Applications}.

Exact approaches like integer linear programming guarantee optimality but scale poorly due to exponential complexity \cite{LP}. Heuristic and metaheuristic methods offer efficient near-optimal solutions but often lack problem-specific knowledge, rely on handcrafted heuristics, and risk premature convergence to local optima \cite{TSP2,Nature-inspired}.

Deep learning has inspired data-driven approaches to COPs such as TSP \cite{MLCO}. Supervised methods train neural networks to imitate solvers for fast inference but depend on labeled solutions, limiting generalization. Reinforcement learning (RL) avoids this by directly optimizing solution quality through interaction, yet RL-based methods still suffer from weak local refinement, poor scalability, and high sample complexity \cite{NCO}.

To address these issues, hybrid frameworks combine neural models with metaheuristics. Non-autoregressive (NAR) “learn-to-predict” methods use neural networks to generate heuristics that guide metaheuristics, rather than directly constructing solutions \cite{DeepACO,RL4CO}. However, current approaches face high variance and low sample efficiency due to simplistic training, while refinement often relies on standard Ant Colony Optimization (ACO) \cite{ACO3}, which exploits TSP structure inefficiently and limits scalability to larger or heterogeneous instances.

\subsection{Paper overview}
We introduce Neural Focused Ant Colony Optimization (NeuFACO), a hybrid framework that integrates deep reinforcement learning with a refined variant of ACO. NeuFACO employs Proximal Policy Optimization (PPO) \cite{PPO} to train a neural policy, offering greater stability and sample efficiency than prior approaches.

For refinement, NeuFACO adapts Focused ACO (FACO) \cite{FACO2023}, an advanced extension that performs targeted modifications around a reference solution. Unlike conventional ACO, which rebuilds full tours each iteration, FACO selectively relocates a few nodes, preserving strong substructures while improving weaker regions. This focused search narrows exploration to promising areas, enhancing efficiency.

By combining neural priors with restricted refinement, NeuFACO balances global guidance and local exploitation. This reduces disruption to near-optimal tours, accelerates convergence, and scales effectively to large TSP instances.

Experiments show that NeuFACO outperforms neural and classical baselines on random TSPs and TSPLIB benchmarks, solving problems with up to 1,500 nodes. Results also confirm the advantage of PPO in producing high-quality heuristic priors, establishing NeuFACO as a robust and generalizable framework for neural-augmented combinatorial optimization.

\section{Related Work}\label{sec:methods}
\subsection{Traveling Salesman Problem}
The Traveling Salesman Problem (TSP) is a classical NP-hard problem in combinatorial optimization \cite{Applications}. It can be formally defined on a complete graph $G=(V,E)$, where the vertex set $V=\{v_1,v_2,\dots,v_n\}$ corresponds to the $n$ cities and the edge set $E$ denotes all pairwise connections among them. Each edge $(i,j)\in E$ has an associated non-negative cost $d_{i,j}$, often representing distance or travel time. For simplicity, this paper focuses on the two-dimensional Euclidean TSP as an illustrative example, where each edge $(i,j)\in E$ has a distance feature
$d_{ij} = \|v_i - v_j\|$.
The task is to determine a Hamiltonian tour $\pi = (\pi_1, \pi_2, \dots, \pi_n)$ that traverses every city once before returning to the origin, with the goal of minimizing the aggregate travel distance or cost
$
C(\pi) = \sum_{i=1}^n d_{\pi_i,\pi_{i+1}},
$
where $\pi$ is a permutation of $V$ representing the visiting order of the cities and $\pi_{n+1}=\pi_1$.

\subsection{Classical Methods}
Classical approaches to the TSP fall into two categories: exact and heuristic-based \cite{TSP2}. Exact methods, such as branch-and-bound \cite{branch} and Integer Linear Programming \cite{LP}, guarantee optimality but suffer factorial time complexity, making them impractical for large instances.

To improve scalability, heuristic and metaheuristic algorithms provide near-optimal solutions in reasonable time \cite{MetaHeuristic}. Local search methods like 2-opt, 3-opt, and Lin–Kernighan \cite{twoopt,LKH} refine tours by replacing subpaths to reduce cost, while metaheuristics such as Ant Colony Optimization \cite{ACO} and Genetic Algorithms \cite{GA} stochastically explore the solution space by simulating natural processes.

\subsection{Ant Colony Optimization}

Ant Colony Optimization is a bio-inspired metaheuristic that emulates the collective foraging behavior of natural ant colonies \cite{ACO}. In the context of the TSP, artificial ants construct solutions by probabilistically choosing the next city according to two key factors: the pheromone trail $\tau_{ij}$, which encodes the learned desirability of selecting edge $(i,j)$, and the heuristic desirability $\eta_{ij}=1/d_{ij}$, which favors shorter edges. The probability $p_{ij}^k$ that ant $k$ moves from city $i$ to city $j$ is typically defined as
\begin{equation}
 p_{ij}^k = \begin{cases}
 \dfrac{\tau_{ij}^\alpha \eta_{ij}^\beta}{\sum_{l\in\calN_i^k} \tau_{il}^\alpha \eta_{il}^\beta} & \text{if } j\in\calN_i^k, \\
 0 & \text{otherwise,}
 \end{cases}
\end{equation}
where $\calN_i^k$ denotes the set of cities not yet visited by ant $k$, and $\alpha$ and $\beta$ determine the relative weighting of pheromone intensity and heuristic desirability. Once all ants have completed their tours, pheromone trails are updated both to reinforce high-quality paths and to allow for evaporation, which encourages exploration:
\begin{equation}
\tau_{ij} \leftarrow (1-\rho)\tau_{ij} + \sum_{k=1}^m \Delta\tau_{ij}^k,
\end{equation}
where $\rho\in(0,1]$ is the evaporation rate, and $\Delta\tau_{ij}^k$ represents the pheromone deposited by ant $k$ proportional to the quality of its solution. Over time, ACO converges toward promising regions of the solution space, but it may encounter slow convergence or premature stagnation.

\begin{figure}[!tb]
  \centering
  \includegraphics[width=\linewidth]{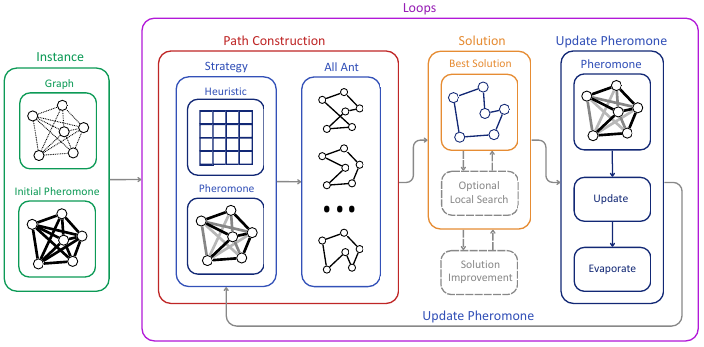}
  \caption{Original Ant Colony Optimization.}
  \label{fig:ACO}
\end{figure}

\subsection{Neural Combinatorial Optimization}
Neural Combinatorial Optimization (NCO) employs deep learning to derive heuristics for NP-hard problems such as the Traveling Salesman Problem (TSP) \cite{MLCO}. Unlike handcrafted approaches, NCO learns directly from data via supervised imitation of high-quality solutions or reinforcement learning on reward signals such as negative tour length. Architectures explored include pointer networks \cite{pointernetwork}, GNNs \cite{gnn}, and more recently Transformers \cite{transformer} and diffusion models \cite{difusco}.

NCO methods are generally categorized as constructive, incrementally generating solutions, or improvement-based, refining initial ones with techniques like 2-opt \cite{NCO}. Non-autoregressive (NAR) variants follow a hybrid paradigm: neural encoders extract heuristic signals that metaheuristics exploit for solution construction, contrasting with end-to-end autoregressive models \cite{RL4CO}.

\begin{figure}[H]
  \centering
  \includegraphics[width=\linewidth]{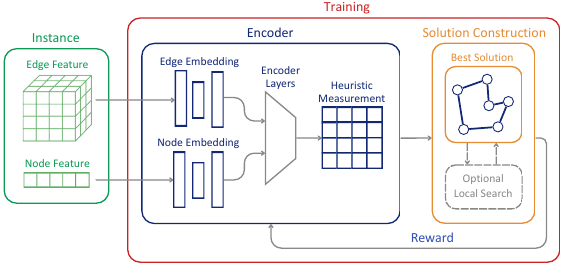}
  \caption{NAR training procedure.}
  \label{fig:DACO}
\end{figure}

In Ant Colony Optimization (ACO), classical heuristics are handcrafted and domain-dependent, limiting scalability and adaptability. NAR methods overcome these issues by learning heuristics from data and embedding them into the ACO pipeline. Early implementations relied on REINFORCE \cite{REINFORCE}, but its high variance and slow convergence hindered performance, motivating more stable reinforcement learning techniques.

\section{Training Methodology}

\subsection{Markov Decision Process (MDP) formulation}
We model TSP solution construction as an MDP. At step $t$, the state $s_t=(\mathcal{X}, S_t, i_t)$ includes the instance, visited set, and current node. The action $a_t\in \mathcal{N}{i_t}\setminus S_t$ is chosen from $\pi\theta(\cdot\mid i_t)$ with feasibility masking. Transitions append $a_t$ to the tour, and the episode ends after all nodes are visited, yielding reward $R=-C(\pi;\mathcal{X})$.

\subsection{Policy Parameterization}
A graph neural network $f_{\theta}$, parameterized by $\theta$, maps graph data $\mathcal{X}$ to:
\begin{itemize}
\item A heuristic matrix $H_\theta \in \RR^{n \times n}$, encoding learned priors over edge transitions.
\item A value estimate $V_\theta(\mathcal{X}) \in \RR$, predicting the expected return of the instance.
\end{itemize}

We train $H_\theta$ as an instance-conditioned heuristic prior using stable on-policy updates. Training is strictly \emph{pheromone-free}: tours are sampled directly from the stochastic policy induced by $H_\theta$, enabling a standard PPO formulation without non-stationary sampler dynamics.
\subsection{Rollouts and Rewards}
For each instance $\mathcal{X}$, we sample a batch of $M$ tours
$\{\pi^{(m)}\}_{m=1}^{M}$ by chaining the policy over the shrinking feasible set:
\[
\pi^{(m)}_{t+1} \sim \pi_\theta(\cdot \mid \pi^{(m)}_t), 
\qquad t=1,\dots,n-1.
\]
Each tour receives a scalar terminal reward equal to the negative tour length:
\begin{equation}
R^{(m)} \;=\; -\, C(\pi^{(m)};\mathcal{X}), 
\qquad m=1,\dots,M.
\end{equation}
We use a per-instance, per-batch baseline $\bar R = \frac{1}{M}\sum_{m=1}^M R^{(m)}$ for variance reduction and to normalize reward and calculated advantage:
\begin{equation}
\tilde R^{(m)} \;=\; R^{(m)} - \bar R, 
\qquad
A^{(m)} \;=\; \tilde R^{(m)} - V_\theta(X).
\end{equation}

\subsection{PPO Objective}
For each sampled path $\pi^{(m)}$, let $\log p_{\theta}(\pi^{(m)})$ denote the log probability of sampling under the current policy, and $\log p_{\theta_{\text{old}}}(\pi^{(m)})$ under the previous policy.
Then the probability ratio is:
\begin{equation}
r^{(m)}(\theta) = \frac{p_{\theta}(\pi^{(m)})}{p_{\theta_{\text{old}}}(\pi^{(m)})}.
\end{equation}
And the clipped PPO surrogate objective is:
\begin{equation}
\begin{aligned}
\mathcal{L}_{\text{policy}}(\theta) = -\frac{1}{M}\sum_{m=1}^M \min\big(&r^{(m)}A^{(m)}, \\
&\text{clip}(r^{(m)},1-\eps,1+\eps)A^{(m)}\big).
\end{aligned}
\end{equation}
The value losses calculated using mean squared error (MSE):
\begin{equation}
\mathcal{L}_{\text{value}}(\theta)
= \frac{1}{M}\sum_{m=1}^{M}\!\left(V_\theta(X) - \tilde R^{(m)}\right)^2
\end{equation}

To promote exploration, entropy is computed on the normalized heuristic distribution: $\tilde{H}_{ij} = H_{ij} / (\sum_k H_{ik})$ and
\begin{equation}
\calH(\tilde{H}) = -\frac{1}n \sum_{i=1}^n\sum_{j=1}^n \tilde{H}_{ij}\log(\tilde{H}_{ij}+\eps).
\end{equation}

Then we define the entropy loss as:
\begin{equation}
\mathcal{L}_{\text{entropy}} = -\beta\calH(\tilde{H}),
\end{equation}

and the final loss to minimize is $\mathcal{L}=\mathcal{L}_{\text{policy}}+\mathcal{L}_{\text{value}}+\mathcal{L}_{\text{entropy}}$.

\section{Solution sampling}
When inference, the ACO constructs a probability distribution over transitions:
$
p_{ij} \propto \tau_{ij}^\alpha \cdot H_{ij}^\beta,
$
where $\tau_{ij}$ is the pheromone and $H_{ij}$ is the learned heuristic, $\alpha$ and $\beta$ are weighting parameters.

\begin{figure}[!htb]
  \centering
  \includegraphics[width=\linewidth]{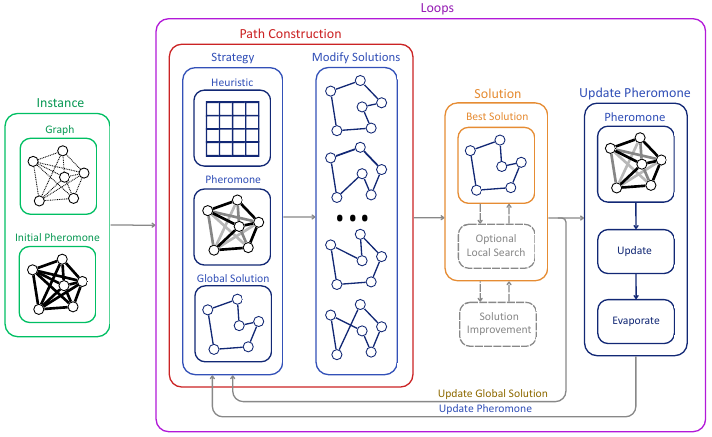}
  \caption{State-of-the-art Ant Colony Optimization.}
  \label{fig:FACO}
\end{figure}

Our method incorporates multiple state-of-the-art enhancements aimed at improving both the computational efficiency and solution quality of ACO algorithms. These improvements address three key aspects of solution construction: efficient node selection, focused modifications to high-quality tours, and scalable local search. Fig.~\ref{fig:FACO} demonstrates the final algorithm for solution sampling.

\subsection{Incorporating the Min--Max Ant System}
To enhance solution quality and stabilize the search process, NeuFACO adopts the Min--Max Ant System (MMAS) \cite{MMAS} for pheromone updates. MMAS modifies the classical Ant System in two key aspects: (1) pheromone trails are restricted to the interval $[\tau_{\min},\tau_{\max}]$ to prevent premature convergence, and (2) updates rely solely on the best-so-far or iteration-best solution, rather than aggregating contributions from all ants.

In our implementation, pheromone updates follow the standard MMAS rule:
\begin{equation}
\tau_{ij} \leftarrow \begin{cases}
(1-\rho)\tau_{ij} + \Delta\tau_{ij}^{\text{best}} & \text{if edge }(i,j) \text{ chosen},\\
(1-\rho)\tau_{ij} & \text{otherwise},
\end{cases}
\end{equation}
\begin{equation}
\tau_{ij} \leftarrow \operatorname{clip}(\tau_{ij},\tau_{\min},\tau_{\max}).
\end{equation}

This mechanism is particularly critical when combined with neural-guided heuristics, as it prevents the learned model from prematurely dominating the search dynamics.

\subsection{Node Selection via Candidate and Backup Lists}
Candidate lists are widely used in ACO to restrict node selection to the $k$-nearest neighbors, thereby reducing complexity. Formally, for current node $i$, the candidate set is
\begin{equation}
\calC_i = \{u_{i,j} \mid j\le k,\; \texttt{visited}(u_{i,j})=0\}.
\end{equation}

When $\calC_i=\varnothing$, classical ACO scans all unvisited nodes, an $O(n)$ operation. To avoid this, we adopt the backup strategy of \cite{ACOTSP-MF}: each node maintains a precomputed backup list $\mathrm{BKP}_i$, from which the first unvisited neighbor is selected. Only if both $\calC_i$ and $\mathrm{BKP}_i$ are empty do we fall back to
\begin{equation}
J = \arg\min_{u:\texttt{visited}(u)=0} d(i,u).
\end{equation}

This hierarchical rule preserves stochastic exploration early on while substantially reducing overhead in the later stages of tour construction.

\subsection{Node Relocation to maintain path structure}
We adopt the focused modification strategy introduced in \cite{FACO2023}, which aims to explore the solution space by making minimal impactful changes to existing high-quality tours. At the start of construction, each ant stochastically copies a complete tour from either the global-best (with probability $p_g$) or the iteration-best candidate and subsequently modifies it by introducing new nodes based on the standard transition rule.

From the selected edge $(u,v)$ and the original tour $\pi$, a relocate move produces a new tour $\pi'$ by removing $v$ from its current position and reinserting it immediately after $u$. Let $p=\operatorname{pred}(v)$ be the predecessor of $v$ in $\pi$, $s=\operatorname{succ}(v)$ be the successor of $v$, and $s_u$ be the successor of $u$. Then the cost change from performing such a move is
\begin{equation}
\Delta C = -d_{p,v} - d_{v,s} - d_{u,s_u} + d_{p,u} + d_{u,v} + d_{v,s_u}.
\end{equation}

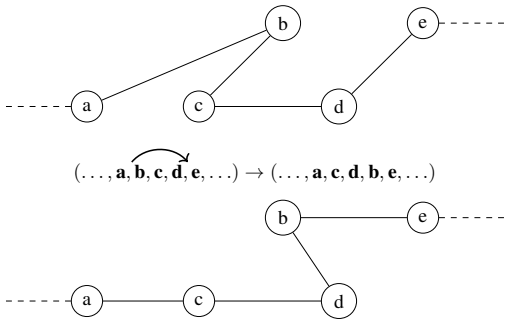
\begin{figure}[ht]
    \centering
    \resizebox{0.85\linewidth}{!}{%

\usetikzlibrary{arrows.meta, positioning, calc}
\begin{tikzpicture}[every node/.style={circle, draw}, node distance=0.5cm, scale=1, transform shape]

\node (a1) at (0,0) {a};
\node (c1) at (2,0) {c};
\node (b1) at (3.5,1.5) {b};
\node (d1) at (4.5,0) {d};
\node (e1) at (6,1.5) {e};

\draw (a1) -- (b1);
\draw (c1) -- (b1);
\draw (c1) -- (d1);
\draw (d1) -- (e1);
\draw[dashed] (a1.west) -- ++(-1.2,0);
\draw[dashed] (e1.east) -- ++(1.2,0);

\node[draw=none, align=center] (permtext) at (3,-1.2) 
    {$(\ldots,\textbf{a},\textbf{b},\textbf{c},\textbf{d},\textbf{e},\ldots) \rightarrow (\ldots,\textbf{a},\textbf{c},\textbf{d},\textbf{b},\textbf{e},\ldots)$};

\draw[->, thick, bend left=50] (0.80,-1) to (1.8,-1);

\node (a2) at (0,-3.5) {a};
\node (c2) at (2,-3.5) {c};
\node (d2) at (4.5,-3.5) {d};
\node (b2) at (3.5,-2) {b};
\node (e2) at (6,-2) {e};

\draw (a2) -- (c2);
\draw (c2) -- (d2);
\draw (d2) -- (b2);
\draw (b2) -- (e2);
\draw[dashed] (a2.west) -- ++(-1.2,0);
\draw[dashed] (e2.east) -- ++(1.2,0);

\end{tikzpicture}

    }
    \caption{Example of a node relocation procedure for edge $(a, b)$.}
    \label{fig:perm-swap}
\end{figure}

To maintain structural similarity to the reference tour, modification is limited: once a predefined minimum number of new edges (MNE) is introduced (we set $\mathrm{MNE}=8$), the ant stops altering the route and continues following the remaining portion of the original tour. This strategy restricts the number of relocated nodes, thereby concentrating the search effort on promising subregions of the solution space while avoiding the computational burden of constructing tours entirely from scratch. Fig. \ref{fig:perm-swap} demonstrates an example of a node relocation operation.

\subsection{Scalable Local Search}
Local search, particularly the 2-opt heuristic, is highly effective for improving ACO-generated solutions but is often applied sparingly due to its computational cost.

To enable frequent refinement, our algorithm incorporates two optimizations from \cite{FACO2023}. First, it tracks modified edges in each constructed tour and applies 2-opt only to those differing from the previously optimized tour. Second, it restricts 2-opt to candidate edges, thereby reducing the search space.

With a constant-size candidate list and a limited set of modified edges, this optimized 2-opt can be applied to every solution efficiently, fully exploiting high-quality tours without prohibitive overhead.

\section{Experiments}
This section presents experimental results to validate the effectiveness of our algorithm. We compare it against other ACO-based methods as well as state-of-the-art neural approaches for solving the TSP. Our source code is built upon codebases of previous ACO-based methods: GFACS (AISTATS 2025) \cite{GFACS}, which itself builds upon DeepACO (NeurIPS 2023) \cite{DeepACO}.

\subsection{Setup}

\subsubsection{Training}
\begin{figure}[H]
  \centering
  \includegraphics[width=\linewidth]{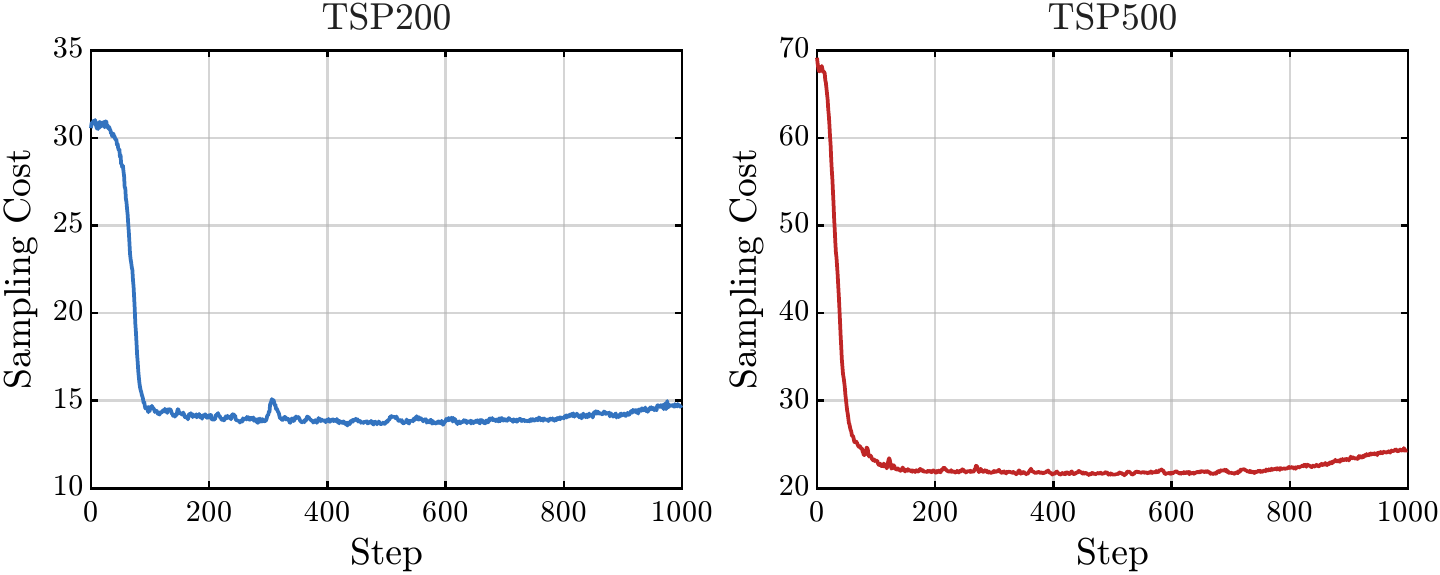}
  \caption{Average objective costs over training step.}
  \label{fig:val}
\end{figure}

We evaluate NeuFACO against DeepACO and GFACS. Training uses 20 PPO steps per epoch, batch size 20, learning rate $\text{lr}=0.001$, clip ratio $\epsilon=0.2$, and entropy coefficient 0.01. Experiments run on an AMD RX6800 GPU and Intel i5-13400 CPU (\SI{2.50}{\giga\hertz}). Fig.~\ref{fig:val} shows test objective cost over training steps, illustrating convergence across instance sizes.

\subsubsection{Benchmark}We evaluate NeuFACO on randomized TSP instances of size 200, 500, and 1000 (128 each) \cite{DeepACO}, and on TSPLib \cite{tsplib} with up to 1500 nodes. Results are averaged over $R=10$ runs for TSPLib and over 128 instances for randomized TSP; baseline results are taken from prior papers. Errors are computed as $E=(\text{cost}-\text{optimal})/\text{optimal}\times100\%$, with optimal values obtained using Concorde \cite{Concorde}.  

Our evaluation consists of two parts: (1) controlled comparison with NAR baselines under identical hyperparameters and hardware, reporting runtime and quality; and (2) literature comparison, compiling solution quality and reported runtimes. As experimental setups vary, runtimes are indicative, while solution quality is emphasized as the primary metric.

\begin{figure*}[!htb]
  \centering

  \resizebox{\textwidth}{!}{%
\begin{tabular}{l cc cc cc}
\toprule[1pt]
& \multicolumn{2}{c}{TSPLib100-299} & \multicolumn{2}{c}{TSPLib300-699} & \multicolumn{2}{c}{TSPLib700-1499} \\
\cmidrule(lr){2-3}\cmidrule(lr){4-5}\cmidrule(lr){6-7}
Method & Gap(\%) & Time & Gap(\%) & Time & Gap(\%) & Time \\
\midrule[1pt]
DeepACO    & 1.50 & 13.70s  & 3.43 & 143.92s  & 4.86 & 1047.59s \\
GFACS      & 1.44 & 13.75s  & 2.46 & 150.49s  & 3.99 & 1141.24s \\
NeuFACO    & \textbf{1.39} & \textbf{0.22s}  & \textbf{2.06} & \textbf{0.48s}  & \textbf{2.98} & \textbf{1.03s} \\
\bottomrule[1pt]
\end{tabular}
\hfill
\begin{tabular}{l cc cc cc}
\toprule[1pt]
& \multicolumn{2}{c}{TSP200} & \multicolumn{2}{c}{TSP500} & \multicolumn{2}{c}{TSP1000} \\
\cmidrule(lr){2-3}\cmidrule(lr){4-5}\cmidrule(lr){6-7}
Method & Gap(\%) & Time & Gap(\%) & Time & Gap(\%) & Time \\
\midrule[1pt]
DeepACO    & 1.89 & 16.86s  & 3.03 & 119.58s  & 4.32 & 659.43s \\
GFACS      & 1.80 & 17.36s  & 1.54 & 136.14s  & 2.48 & 730.56s \\
NeuFACO    & \textbf{1.66} & \textbf{0.28s}  & \textbf{1.50} & \textbf{0.74s}  & \textbf{2.00} & \textbf{1.98s} \\
\bottomrule[1pt]
\end{tabular}
  }

  \caption{Comparison of DeepACO, GFACS and NeuFACO priors. All models are retrained with their recommended parameters, using the same number of epochs and evaluation settings.}
  \label{fig:NAR}
\end{figure*}

\subsection{Comparison with Other NAR Solvers}

\begin{figure}[htbp]
  \centering
  \includegraphics[width=\linewidth]{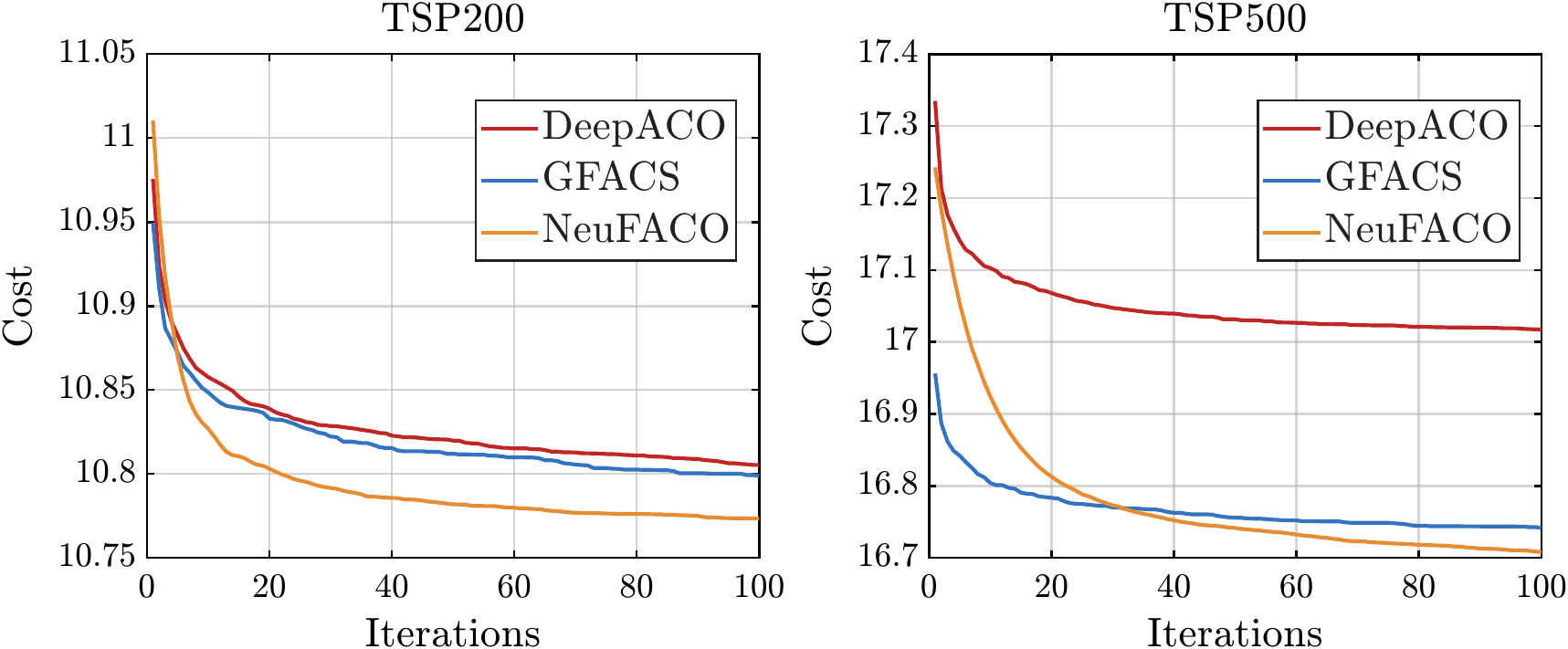}
  \caption{Objective cost over ACO iterations between different priors.}
  \label{fig:over_iterations}
\end{figure}

We compare NeuFACO to two non-autoregressive baselines, DeepACO and GFACS, both of which outperform classical ACO without neural guidance \cite{DeepACO,GFACS}. For all NAR models we fix the number of ants $M=100$, run $I=100$ ACO iterations, $\alpha = 1$, $\beta = 1$, and an evaporation rate of $\rho=0.1$. Unlike DeepACO and GFACS, which employ neural-guided perturbation, NeuFACO applies scalable local search at every iteration, providing effective refinement without additional perturbation mechanisms. NeuFACO-specific hyperparameters are: global path probability $p_g=0.01$, minimum new edges per iteration $\mathrm{MNE}=8$, candidate list size $k=20$, backup list size $\mathrm{BKP}=64$,
$$
\tau_{\max}=\frac{1}{(1-\rho)\,g_b},
\qquad
\tau_{\min}=\min\!\left(\tau_{\max},\;\frac{\tau_{\max}\,(1-p_{\text{best}}^{1/k})}{(k-1)\,p_{\text{best}}^{1/k}}\right),
$$
with $g_b$ denoting the global-best objective and $p_{\text{best}}=0.1$ as suggested in \cite{FACO2023}.

\begin{figure}[h]
  \centering
  \includegraphics[width=\linewidth]{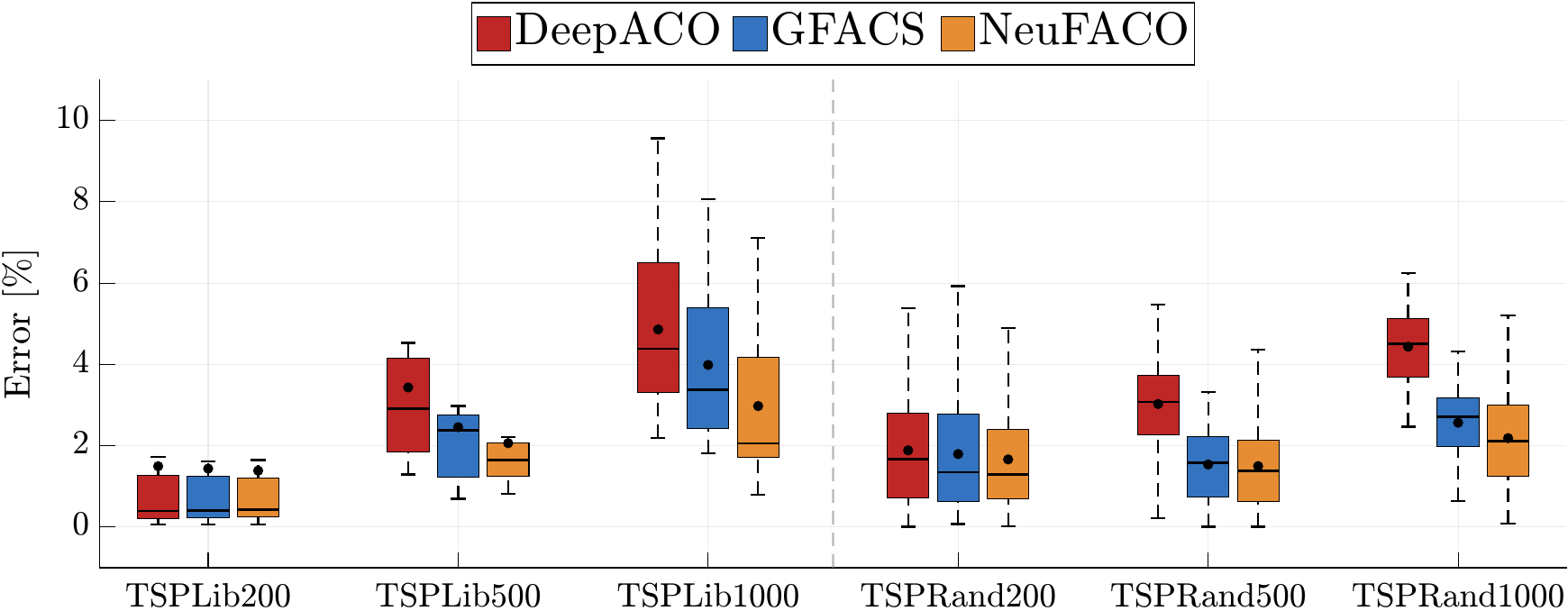}
  \caption{Minimum, maximum and average errors comparison.}
  \label{fig:candle}
\end{figure}

Fig.~\ref{fig:NAR} shows that NeuFACO achieves the highest solution quality among DeepACO, GFACS, and itself across nearly all datasets. Under identical settings, NeuFACO attains up to ~60$\times$ lower wall-clock time than prior NAR methods, thanks to a refined sampler that exploits neural heuristics with minimal quality loss. PPO with entropy regularization further mitigates policy degradation, ensuring solution quality comparable to or exceeding existing models. Gains are most pronounced on large instances, such as TSPLib graphs with up to 1500 nodes, highlighting NeuFACO’s efficiency on large-scale TSPs.

Fig.~\ref{fig:over_iterations} confirms faster convergence of average objective costs, while Fig.~\ref{fig:candle} shows consistently lower variance and superior performance across instance sizes.

\subsection{Comparison with other RL solvers}
In this section, we compare NeuFACO with problem-specific RL approaches for the TSP, including established baselines (AM \cite{transformer}, POMO \cite{pomo}, Pointerformer \cite{pointerformer}), heatmap-based DIMES \cite{dimes}, improvement-based SO \cite{SelectAndOptimize}, and more recent methods such as H-TSP \cite{htsp}, GLOP \cite{glop}, INViT \cite{invit}, LEHD \cite{lehd}, BQ \cite{bq}, SIGD \cite{sigd}, and SIT \cite{sit}. NeuFACO is run with $M=256$ ants and $I=1000$ iterations to leverage its efficient sampling, and reported times include both heuristic inference and solution sampling.


\begin{figure}[!tb]
\centering
\small 
\setlength{\tabcolsep}{0pt} 
\renewcommand{\arraystretch}{1.05} 

\footnotesize Results for other methods are taken from \cite{GFACS,glop,htsp}. For NeuFACO, we reported the averaged objective values as well as both heuristic inference and solution sampling runtimes. \\[3pt]
\resizebox{\columnwidth}{!}{%
\begin{tabular}{l ccc ccc}
\toprule[1pt]
& \multicolumn{3}{c}{TSP500} & \multicolumn{3}{c}{TSP1000}\\
\cmidrule(lr){2-4}\cmidrule(lr){5-7}
Method & Obj. & Gap(\%) & Time & Obj. & Gap(\%) & Time \\
\midrule[1pt]
Concorde           & 16.55 & --    & 10.7s & 23.12 & --   & 108s \\
\cmidrule(lr){1-1}\cmidrule(lr){2-4}\cmidrule(lr){5-7}
AM                 & 21.46 & 29.07 & 0.6s & 33.55 & 45.10 & 1.5s \\
POMO          & 20.57 & 24.40 & 0.6s & 32.90 & 42.30 & 4.1s \\
DIMES              & 17.01 & 2.78  & 11s  & 24.45 & 5.75  & 32s \\
SO                 & 16.94 & 2.40  & 15s  & 23.77 & 2.80  & 26s \\
Pointerformer      & 17.14 & 3.56  & 14s  & 24.80 & 7.90  & 40s \\
\cmidrule(lr){1-1}\cmidrule(lr){2-4}\cmidrule(lr){5-7}
H-TSP (AAAI'23)             & 17.55 & 6.22  & 0.18s  & 24.65 & 6.62  & 0.33s\\
GLOP  (AAAI'24)             & 16.91 & 1.99 & 1.5m  & 23.84 & 3.11  & 3.0m\\
INViT-3V+Greedy (ICML'24)    & 16.78 & 1.56 & 5.48s & 24.66 & 6.66  & 9.0s \\
LEHD+Greedy (NeurIPS'23)             & 16.78 & 1.56 & 0.13s & 23.84 & 3.11 & 0.8s \\
BQ+Greedy (NeurIPS'23)             & 16.72 & 1.18 & 0.36s & 23.65 & 2.30  & 0.9s\\
SIGD+Greedy (NeurIPS'23)             & 16.71 & 1.17 & 0.23s & 23.57 & 1.96  & 1.2s\\
SIT+Greedy (NeurIPS'25)                   & 16.7 & 1.08 & 14.95s & 23.57 & 1.95 & 0.2s \\
\cmidrule(lr){1-1}\cmidrule(lr){2-4}\cmidrule(lr){5-7}
DeepACO (NeurIPS'23)            & 16.84 & 1.77 & 15s & 23.78 & 2.87 & 1.1m \\
GFACS (AISTATS'25)             & 16.80 & 1.56 & 15s & 23.72 & 2.63 & 1.1m \\
\cmidrule(lr){1-1}\cmidrule(lr){2-4}\cmidrule(lr){5-7}
NeuFACO            & 16.43 & 1.33 & 0.91s & 23.34 & 1.16 & 0.98s \\
            &  &  & +14.95s &  &  & +40.03s \\

\bottomrule[1pt]
\end{tabular}
}

\caption{Broader literature comparisons on TSP500 and TSP1000.}
\label{fig:tsp-results}
\end{figure}

Results demonstrate that NeuFACO consistently achieves superior or highly competitive performance across all baselines. It surpasses every established methods and remains competitive with recently proposed approaches, often delivering better solution quality at the cost of slightly longer runtimes. The experiments also highlight a key bottleneck of NAR methods, including ours: solution sampling remains CPU-bound, leading to higher runtime despite comparable amortized inference. Nevertheless, NeuFACO marks a significant advance for both NAR and RL-based approaches to the TSP.

\section{Conclusion}
In conclusion, NeuFACO integrates deep reinforcement learning with a refined ACO framework to address limitations of non-autoregressive models for the TSP. By combining PPO-based policy learning with targeted refinement around high-quality solutions, it achieves a strong balance between global guidance and local exploitation. This synergy preserves solution structure, accelerates convergence, and scales effectively to large instances. Experiments show that NeuFACO consistently achieves superior or highly competitive performance compared to a wide range of neural baselines on both randomized and benchmark datasets. While runtimes are longer due to CPU-bound sampling, the method delivers higher solution quality, underscoring the effectiveness of PPO-guided priors in enhancing ACO and establishing NeuFACO as a robust and generalizable framework for neural-augmented combinatorial optimization.


\bibliographystyle{IEEEtran}
\bibliography{refs}

\end{document}